\documentclass[letterpaper, 10 pt, journal, twoside]{IEEEtran}
\usepackage[letterpaper,
            left=0.75in, right=0.75in,
            top=0.75in, bottom=0.75in]{geometry}
\usepackage{bm}
\usepackage{cite}
\usepackage{amsmath,amssymb,amsfonts}
\usepackage{algorithm}
\usepackage{algorithmic}
\usepackage{graphicx}
\usepackage{caption}
\usepackage{booktabs}
\usepackage{pifont}
\usepackage{url}
\usepackage{xcolor}
\usepackage{multirow}

\newcommand{\SEthree}{\mathrm{SE}(3)}
\newcommand{\SOthree}{\mathrm{SO}(3)}
\newcommand{\sethree}{\mathfrak{se}(3)}

\newcommand{\R}{\mathbb{R}}
\newcommand{\bT}{\mathbf{T}}
\newcommand{\bR}{\mathbf{R}}
\newcommand{\bc}{\mathbf{c}}
\newcommand{\bp}{\mathbf{p}}
\newcommand{\bv}{\mathbf{v}}

\newcommand{\bE}{\mathbf{E}}
\newcommand{\bI}{\mathbf{I}}
\newcommand{\bG}{\mathbf{G}}
\newcommand{\bW}{\mathbf{W}}

\newcommand{\bSig}{\boldsymbol{\Sigma}}
\newcommand{\bxi}{\boldsymbol{\xi}}
\newcommand{\bx}{\boldsymbol{x}}
\newcommand{\bom}{\boldsymbol{\omega}}
\newcommand{\Log}{\operatorname{Log}}

\usepackage{colortbl}

\definecolor{bestblue}{HTML}{DCEAF7}
\definecolor{secondorange}{HTML}{FCEBD5}
\definecolor{rankone}{HTML}{9EC5E8}
\definecolor{ranktwo}{HTML}{C9DFF2}
\definecolor{rankthree}{HTML}{EAF3FB}

\begin{document}
\bstctlcite{IEEEexample:BSTcontrol}

\title{\rule{0pt}{0.4in}Shades of Success: Evaluating Manipulation Quality from Object-Centric Constraints}

\author{
    Lorena Maria Genua$^{1}$, 
    Sarvesh Prajapati$^{1}$, 
    Ananya Trivedi$^{1}$,
    Ta\c{s}k{\i}n Pad{\i}r$^{1\ddagger}$
    \thanks{$^{1}$Institute for Experiential Robotics, Northeastern    University, Boston, Massachusetts, USA.
    {\tt\small \{genua.l, prajapati.s, trivedi.ana, t.padir\}@northeastern.edu}}
    \thanks{$^\ddagger$Ta\c{s}k{\i}n Pad{\i}r holds concurrent appointments as a Professor of Electrical and Computer Engineering at Northeastern University and as an Amazon Scholar. This paper describes work performed at Northeastern University and is not associated with Amazon.}
}

\makeatletter
\let\@oldmaketitle\@maketitle
\renewcommand{\@maketitle}{\@oldmaketitle
    \vspace{1ex}
    \begin{center}
        \setcounter{figure}{0}
        \includegraphics[width=\linewidth]{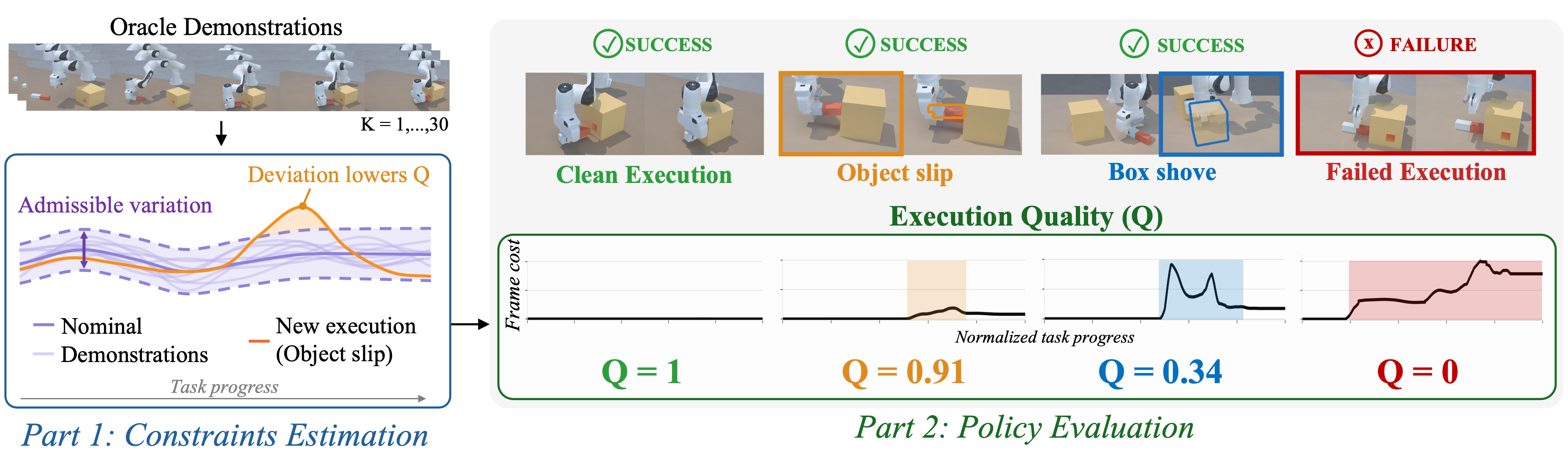}
        \captionof{figure}{To capture subtle variations in execution quality that binary success labels miss, our proposed framework SPECTRA operates in two stages: \textbf{(Part 1)}~from a small set of successful oracle demonstrations, it learns which object relationships are constrained at each stage of the task, and how tightly; and \textbf{(Part 2)}~for an unseen execution, deviations that exceed these tolerances increase the frame cost and lower the quality index $Q$.}
        \label{fig:teaser}
        \vspace{-5ex}
    \end{center}
}
\makeatother

\maketitle

\begin{abstract}
As learned manipulation policies become increasingly capable of solving practical manipulation tasks, assessing \textit{how well} a task is performed becomes a fundamental challenge.
While binary labels confirm task success, they cannot distinguish successful executions with different levels of geometric precision.
To address this, we introduce SPECTRA, an object-centric framework for quantifying execution quality in manipulation of rigid objects.
From a small set of oracle demonstrations generated in simulation, SPECTRA represents object interactions through progress-aligned relative poses.
SPECTRA identifies which relative-pose directions are constrained and their admissible variations from demonstrations.
These constraints establish a reference for evaluating robotic manipulation tasks.
To score an unseen execution, deviations beyond the admissible variation are summarized by a Task Quality Index $Q \in [0,1]$.
Across seven simulated tasks, SPECTRA recovers constraints with 94\% accuracy using 30 demonstrations per task.
On rollouts from three learned policies, its scores achieve pooled Spearman correlations with human judgments of $\rho=0.81$ over all rollouts and $\rho=0.61$ among successful rollouts, exceeding all baselines in both subsets.
Using the task reference learned in simulation, we use SPECTRA to select the highest-scoring demonstrations for training a diffusion policy on hardware.
Exploratory results show higher success rate than training on randomly selected demonstrations.

\noindent Project page: \url{https://spectra-manip.github.io/}

\vspace{-2ex}
\end{abstract}

\IEEEpeerreviewmaketitle

\section{Introduction}
\label{sec:intro}

Imitation learning from human demonstrations has extended robot manipulation capabilities to complex, contact-rich tasks~\cite{black2025pi05, chi2025diffusion, zhao2023act, gao2023kvil, hsu2025spot}.
As these systems become increasingly capable, how their performance is evaluated becomes an important consideration. Common benchmarks rely on a binary notion of success~\cite{liu2023libero,tao2024maniskill3}: an execution either successfully completes the task or it does not.
%
% Although this establishes whether a policy can complete a task,
However, this provides limited information about \textit{how well} the task was performed.

A successful manipulation task can vary substantially in how it is executed. 
Consider a robotic assembly task such as peg insertion: one execution may maintain the peg's alignment with the hole, while another may introduce deviations that push the box out of position during alignment (Fig.~\ref{fig:teaser}). 
Success rate does not capture such differences, and therefore cannot distinguish policies that reach the same goal with different levels of geometric precision during execution.
This motivates a fundamental question: \textit{how should the quality of a successful manipulation task be measured?}

Although recent work has moved beyond binary task success, manipulation quality is still rarely captured by a single task-specific measure.
Existing evaluations often decompose quality into multiple separate
indicators~\cite{wang2025roboeval, liu2026evalactions, valle2025evaluating, yang2026robolab},
or rely on manually specified criteria~\cite{mao2026beyond} and human
judgments~\cite{liu2026evalactions, atreya2025roboarena}.
Others use learned evaluators that estimate task
progress~\cite{liu2026prmasajudge15toolkitrobot, liang2026robometer} or compare rollout videos with vision-language models~\cite{wang2026r2seval}.
While these approaches provide richer feedback than binary outcomes, evaluating manipulation through task-specific geometric relationships remains underexplored.

Successful demonstrations could offer a complementary basis for defining execution quality, as they encode examples of how a task is expected to be performed~\cite{darpino2017clearn}. 
Although individual trajectories may differ, they preserve common geometric relationships between the objects involved~\cite{gao2023kvil,hsu2025spot}, while their variability provides evidence of where greater precision is required~\cite{gao2023kvil}.
Prior work has used these regularities to learn task constraints and guide execution~\cite{mronga2021learning,overbeek2025versatile}.
We build on this idea to establish a reference for evaluating geometric execution quality: which relationships should an execution preserve, and how much deviation is admissible?
This reference must account for the changing requirements of the task, since motion that varies freely during approach may become tightly constrained during insertion~\cite{das2026learning}.

In this work, we introduce SPECTRA, an object-centric framework for evaluating geometric execution quality in rigid-body manipulation.
From successful oracle demonstrations generated in simulation, SPECTRA learns task-specific geometric constraints that serve as a reference for evaluating unseen executions.
It aligns demonstrations by task progress and, at each stage, eigendecomposes the covariance of relative-pose deviations across demonstrations to identify constrained directions and their admissible variation.
To score an unseen execution, SPECTRA penalizes deviations along constrained directions and aggregates the resulting costs into a scalar quality index~$Q$.
SPECTRA learns its task-specific reference from as few as 30 demonstrations, without human labels.
Because the formulation relies on relative object and gripper poses, it is independent of the arm’s kinematic structure and policy architecture.

Across seven simulated manipulation tasks, we demonstrate that SPECTRA recovers expected constraint counts, distinguishes successful executions with different levels of geometric precision, and agrees with human judgments closely than all baselines, both over all rollouts and among successful executions.
Hardware experiments further demonstrate that the reference learned in simulation transfers to scoring hardware executions.
Finally, we present an exploratory study showing how SPECTRA quality scores can support demonstration selection for policy learning.

% The main contributions of this paper are:

% \begin{itemize}
%     \item SPECTRA, a framework that scores the quality of a rigid-body manipulation execution, capturing differences in geometric precision that binary success labels miss.

%     \item An evaluation across seven manipulation tasks, examining constraint recovery, sensitivity to controlled deviations, and agreement with human judgments of policy rollouts.

%     \item A hardware evaluation demonstrating that simulation-learned constraints transfer to scoring hardware executions, together with an exploratory study of SPECTRA-guided demonstration selection for policy learning.
% \end{itemize}

\section{Related Work}
\label{sec:related}
\subsection{Manipulation Performance Evaluation}

Recent work recognizes that binary task success provides only a limited view of manipulation performance.
One line of work augments task completion with additional execution metrics, such as completion time, trajectory length, smoothness, jerk, collisions, and object slips~\cite{valle2025evaluating,wang2025roboeval, yang2026robolab}. 
RoboLab~\cite{yang2026robolab} additionally uses spectral arc length to characterize motion smoothness.
These indicators capture complementary aspects of execution, but combining them requires choices about normalization and weighting.
Other approaches introduce task-dependent specifications of desirable execution. 
For instance, LIBERO-Elegant~\cite{mao2026beyond} evaluates requirements such as target pose accuracy, task sequencing, and collision avoidance. 
Its evaluation criteria are manually specified, and constraint satisfaction is annotated with binary labels that supervise a learned Elegance Critic.
Human preferences provide another basis for evaluation. RoboArena~\cite{atreya2025roboarena} aggregates pairwise human judgments across tasks and environments to rank policies.
Eval-Actions~\cite{liu2026evalactions} combines expert grades, preference rankings, and calibrated motion indicators, and introduces AutoEval to predict execution quality from video and kinematic summaries.
These approaches capture broader judgments of desirable behavior but depend on human feedback.
R2S-Eval~\cite{wang2026r2seval} instead uses a vision-language model to compare rollout videos and aggregates its preferences into policy rankings.
Progress-based approaches assess execution through advancement toward the task goal. PRM-as-a-Judge 1.5~\cite{liu2026prmasajudge15toolkitrobot} derives measures of progress efficiency, regression, stagnation, and recovery from learned progress curves. Robometer~\cite{liang2026robometer} combines frame-level progress supervision with trajectory preferences and applies its learned rewards to policy learning and demonstration filtering.
However, these measures characterize how an execution advances toward the goal, rather than how it is carried out along the way.

Together, these methods highlight a shift beyond binary completion, but leave a gap for task-specific evaluation of geometric precision without manually crafted constraints or human quality labels.
With SPECTRA, we make the quality of object interactions an explicit, graded, and interpretable measure of manipulation performance.

\subsection{Learning Task Constraints from Demonstrations}

Demonstrations of the same task rarely follow identical trajectories, yet they consistently preserve the aspects of the task that matter.
Constraint-learning approaches exploit this regularity to recover task constraints rather than a single trajectory to reproduce.
Consistency across demonstrations reveals which variables are relevant to a task, and how their importance changes as it unfolds~\cite{ureche2015task,bowen2015asymptotically}.
Geometric representations then express these variables as relationships between objects and keypoints, distinguishing constrained and permitted motion~\cite{cai2020inferring,gao2023kvil}.
Demonstrated variability is also used to determine how tightly execution should follow a learned reference~\cite{bowen2015asymptotically}.
In these works, however, learned constraints serve to generate or regulate robot motion.
SPECTRA instead assesses executions produced by a given planner or policy, using the learned constraints to quantify geometric execution quality and identify deviations along task-relevant directions.
    
\section{Methodology}
\label{sec:method}
\begin{figure*}[t!]
    \centering
    \includegraphics[width=\linewidth]{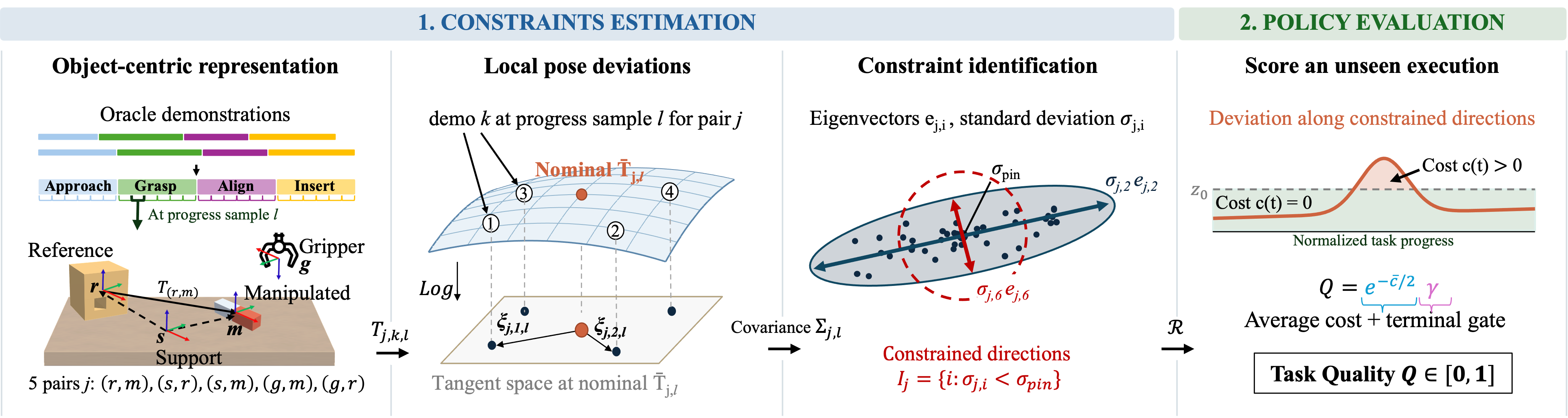} 
    \caption{Overview of SPECTRA. \textbf{1.~Constraints estimation.} Oracle demonstrations are segmented and aligned to shared progress samples. At each progress sample $l$, the task is represented by the relative poses between the reference object $r$, the manipulated object $m$, the gripper $g$, and the support object $s$. Deviations from the nominal pose $\bar{\bT}_{j,l}$ are mapped to its tangent space, and eigen decomposition of their covariance $\bSig_{j,l}$ identifies the constrained directions $\mathcal{I}_j$, whose standard deviation falls below $\sigma_{\mathrm{pin}}$. \textbf{2.~Policy evaluation.} An unseen execution incurs cost only where its normalized deviation along constrained directions exceeds the tolerance $z_0$. The average cost and a terminal gate give the \textbf{Task Quality Index $Q\in[0,1]$}.}    
    \label{fig:method}
    \vspace{-1ex}
\end{figure*}

SPECTRA builds a \emph{task-specific, progress-dependent} representation of geometric constraints from simulated oracle demonstrations.
%
% The premise is that successful executions do not need to reproduce the same trajectory: at any stage of the task, some directions of motion vary
% substantially across demonstrations while others remain tightly regulated.
%
The method proceeds in two parts (Fig.~\ref{fig:method}).
In Part~1 (Sec.~\ref{sec:constraints_estimation}), we learn a task representation~$\mathcal{R}$, specifying at each stage of the task which relative-pose directions are constrained
and how much deviation each admits.
In Part~2 (Sec.~\ref{sec:policy_scoring}), we use $\mathcal{R}$ to evaluate an unseen execution, aggregating deviations beyond the calibrated tolerances into a Task Quality Index~$Q \in [0,1]$.

\subsection{Part 1: Constraints Estimation}
\label{sec:constraints_estimation}

\subsubsection{\textbf{Object-Centric Task Representation}}
\label{subsec:object_representation}

For each task, we generate $K$=30 successful oracle demonstrations using scripted solvers built on ManiSkill’s~\cite{tao2024maniskill3} motion planner with Franka Panda Arm, with randomized initial object positions and orientations.
We additionally vary free-space transport motions to introduce trajectory diversity while preserving critical sub-tasks, such as grasping, insertion, and placement.
We use simulation because it provides a controlled reference with precise object poses, avoiding operator-dependent variability in teleoperation and pose-estimation errors in hardware data collection.

Because demonstrations differ in duration and execution speed, we align them by task progress rather than by time.
A vision-language model defines an ordered sequence of $P$ semantic sub-tasks (e.g. approach, grasp, reach, release) from the task instruction and scene context, and pose-based heuristics identify the frames at which each demonstration transitions between them.
Within each sub-task, elapsed time is normalized to $[0,1]$ and divided into $M$ equal-width progress bins, yielding $L=PM$ canonical progress samples across the full task.
A given bin thus refers to the same stage of the task in every demonstration, regardless of how long that sub-task took (Fig.~\ref{fig:method}).

After this common progress axis between demonstrations is established, SPECTRA characterizes manipulation through the relative poses of task-relevant bodies at corresponding stages of the task.
The reference object $r$ is assigned as the non-support task object with the lowest temporal centroid variability, averaged across demonstrations.
The manipulated objects are those moved by the robot during the task.
Each manipulated object $m$ is represented relative to $r$.
Both the manipulated objects and the reference object are additionally represented relative to the supporting object $s$, such as a table.
The gripper $g$ is represented by its tool-center-point (TCP) frame,
relative to which the manipulated and reference objects are expressed.

Let $\mathcal{C}$ denote the set of ordered interaction pairs defined above. At frame $t$ of demonstration $k \in \{1,\ldots,K\}$,
the pose of object $b$ relative to object $a$, for
$j=(a,b)\in\mathcal{C}$, is
\begin{equation}
    \bT_{j,k}(t)
    =
    \bT_{a,k}(t)^{-1}\bT_{b,k}(t)
    \in\SEthree,
    \label{eq:relative_pose}
\end{equation}
where $\bT_{a,k}(t)$ and $\bT_{b,k}(t)$ denote the poses of objects $a$ and $b$ in the camera frame, respectively.

At each canonical progress sample $l \in \{1,\ldots,L\}$, demonstration $k$ contributes one representative relative pose $\bT_{j,k,l}$ for each interaction pair $j \in \mathcal{C}$.
When multiple valid frames fall within the same bin, their Lie-group mean is used. 
%
% This gives each demonstration equal weight at each progress sample.

\subsubsection{\textbf{Local Pose Deviations}}
\label{subsec:nominal_configuration}

To characterize variation across demonstrations, we express each relative pose as a local deviation from a nominal configuration.
For interaction pair $j$ and progress sample $l$, the deviation of
demonstration $k$ from the nominal pose $\bar{\bT}_{j,l}\in\SEthree$ is
\begin{equation}
    \bxi_{j,k,l}
    =
    \Log\!\left(\bar{\bT}_{j,l}^{-1}\bT_{j,k,l}\right)
    =
    \begin{bmatrix}\bom_{j,k,l}\\ \bv_{j,k,l}\end{bmatrix}
    \in \sethree \simeq \R^6,
    \label{eq:tangent_deviation}
\end{equation}
where $\bar{\bT}_{j,l}$ is the Lie-group mean, defined implicitly by $\sum_{k=1}^{K}\bxi_{j,k,l}=\mathbf{0}$ and computed by iterative
averaging in the tangent space.
For $\bar{\bT}_{j,l}=(\bar{\bR},\bar{\bc})$ and $\bT_{j,k,l}=(\bR,\bc)$,
\begin{equation}
    \bom=\Log\!\left(\bar{\bR}^{-1}\bR\right),
    \qquad
    \bv=\mathbf{V}(\bom)^{-1}\bar{\bR}^{-1}(\bc-\bar{\bc}),
    \label{eq:se3_components}
\end{equation}
where $\mathbf{V}(\bom)$ is the $3\times3$ left Jacobian of $\SOthree$.
Each pair at each progress sample thus has its own local tangent space, centered at its nominal configuration, in which cross-demonstration variation is characterized in $\R^6$.

Because $\bom$ and $\bv$ have different units, their variability is not
directly comparable.
Following object-displacement metrics~\cite{bregier2018defining},
we measure a deviation by the first-order root mean square displacement it induces on keypoints of the measured object.
For each pair $j$, let $\mathcal{K}_j\subset\R^3$ be six non-collinear keypoints obtained by farthest-point sampling of object $b$'s observed surface points, expressed in its body frame.
For a deviation $\bxi_{j,k,l}$, the first-order displacement of a
keypoint $\bp\in\mathcal{K}_j$, expressed in the nominal body frame, is
\begin{equation}
    \delta\bp
    =
    \bom\times\bp+\bv
    =
    \begin{bmatrix}-\hat{\bp} & \bI_3\end{bmatrix}\bxi,
    \label{eq:keypoint_displacement}
\end{equation}
where $\hat{\bp}$ is the skew-symmetric matrix with
$\hat{\bp}\bx=\bp\times\bx$.
Averaging over the keypoints gives
\begin{equation}
    d_j^2(\bxi)
    =
    \frac{1}{|\mathcal{K}_j|}\sum_{\bp\in\mathcal{K}_j}\|\delta\bp\|_2^2
    =
    \bxi^\top\bG_j\bxi,
    \label{eq:physical_distance}
\end{equation}
with
\begin{equation}
    \bG_j
    =
    \frac{1}{|\mathcal{K}_j|}\sum_{\bp\in\mathcal{K}_j}
    \begin{bmatrix}
        -\hat{\bp}^{\,2} & \hat{\bp}\\
        -\hat{\bp} & \bI_3
    \end{bmatrix},
    \label{eq:metric_tensor}
\end{equation}
which depends on the geometry of the object $b$ in pair
$j=(a,b)$ and is positive definite because the keypoints are
non-collinear.
We then work in the scaled coordinates
\begin{equation}
    \tilde{\bxi}_{j,k,l}=\bW_j\bxi_{j,k,l},
    \qquad
    \bW_j=\bG_j^{1/2},
    \label{eq:metric_coordinates}
\end{equation}
so that, under a first-order approximation, $\|\tilde{\bxi}\|_2$ is the
root mean square keypoint displacement in meters.

\subsubsection{\textbf{Constraint Identification}}
\label{sec:constraints}

We identify the constrained directions of each relation through a spectral analysis of scaled deviations across demonstrations.
For pair $j$ and progress sample $l$, we compute the cross-demonstration
covariance and its eigendecomposition,
\begin{equation}
    \bSig_{j,l}
    =
    \frac{1}{K-1}
    \sum_{k=1}^{K}
    \tilde{\bxi}_{j,k,l}\tilde{\bxi}_{j,k,l}^{\top}
    =
    \bE_{j,l}\boldsymbol{\Lambda}_{j,l}\bE_{j,l}^{\top},
    \label{eq:sample_cov}
\end{equation}
where the columns $\bE_{j,i}(l)$ of $\bE_{j,l}$ are the principal
directions and
$\boldsymbol{\Lambda}_{j,l}=\operatorname{diag}(\nu_{j,1}(l),\ldots,\nu_{j,6}(l))$
contains their variances.
To ensure positive definiteness and bound the normalization below, the
eigenvalues are floored at $\sigma_0^2$, with $\sigma_0=3\,\mathrm{mm}$, and the standard
deviations $\sigma_{j,i}(l)=\sqrt{\nu_{j,i}(l)}$ are computed from the
floored values.
Each principal direction is a combination of rotation and translation,
and its standard deviation measures, in meters, how much the
demonstrations vary along it.

A direction is classified as constrained when its standard deviation
falls below a fixed threshold $\sigma_{\mathrm{pin}}$:
\begin{equation}
    \mathcal{I}_{j}(l)
    =
    \left\{
        i : \sigma_{j,i}(l)<\sigma_{\mathrm{pin}}
    \right\}.
    \label{eq:constraint_set}
\end{equation}
A constrained direction is thus one that successful demonstrations
consistently regulate, such as the alignment of a peg with its hole,
while directions with larger variation, such as the yaw of a cube
placed on a target, are left unpenalized.
The number of constrained directions $|\mathcal{I}_j(l)|$ can change
along the task, capturing how its requirements tighten or relax.
We further split $\mathcal{I}_j(l)$ into directions with a measured
spread, $\mathcal{M}_j(l)=\{i\in\mathcal{I}_j(l):\sigma_{j,i}(l)>\sigma_0\}$,
and floored directions, $\mathcal{F}_j(l)=\mathcal{I}_j(l)\setminus\mathcal{M}_j(l)$,
along which the demonstrations agree to within the floor.

% To estimate how much deviation is admissible along constrained directions, we normalize deviations by their demonstrated spread.
Because the admissible deviation along a direction depends on how tightly the demonstrations regulate it, we normalize deviations by their demonstrated spread, so that a single tolerance applies across directions, pairs and progress samples.
For a frame $t$ of a demonstration, we define the normalized deviation along each constrained direction as
\begin{equation}
    z_{j,i}(t)
    =
    \frac{\bE_{j,i}(t)^\top\tilde{\bxi}_{j}(t)}{\sigma_{j,i}(t)},
    \qquad i\in\mathcal{I}_j(t).
    \label{eq:demonstration_normalization}
\end{equation}
The task-level tolerance $z_0$ is the 95th percentile of $|z_{j,i}(t)|$ over all frames of the $K$ demonstrations, all pairs, and all measured directions $i\in\mathcal{M}_j(t)$.
Along floored directions, the normalization scale is set by the prescribed floor $\sigma_0$ rather than the demonstrated spread. We therefore estimate $z_0$ using only measured directions, whose normalization reflects the observed variability across demonstrations.

The resulting task representation is
\begin{equation}
    \mathcal{R}
    =
    \left\{
        \bar{\bT}_{j,l},\,
        \bSig_{j,l},\,
        \mathcal{I}_{j}(l),\,
        z_0
    \right\}.
    \label{eq:task_representation}
\end{equation}

\subsection{Part 2: Policy Evaluation}
\label{sec:policy_scoring}

\subsubsection{\textbf{Progress-Aligned Execution Scoring}}
\label{subsec:scoring}

An unseen execution is segmented into the same semantic sub-tasks using
pose-based transition heuristics, and elapsed time within each sub-task
is normalized to obtain its progress $l(t)$, as in
Sec.~\ref{subsec:object_representation}.
At this progress, we interpolate the neighboring reference samples
within the sub-task, using SLERP for rotation, linear interpolation for
translation, and Log-Euclidean interpolation for covariance.
For each interaction pair $j$, the scaled deviation $\tilde{\bxi}_j(t)$
from the interpolated nominal pose follows from~\eqref{eq:tangent_deviation}
and~\eqref{eq:metric_coordinates}, and the interpolated covariance gives
the principal directions $\bE_{j,i}(t)$, standard deviations
$\sigma_{j,i}(t)$, and constrained directions
$\mathcal{I}_j(t)=\mathcal{M}_j(t)\cup\mathcal{F}_j(t)$.
Deviations along constrained directions are normalized as
in~\eqref{eq:demonstration_normalization}.
To score an execution, we first summarize its deviations at each frame into a single frame cost $c(t)$.
Each normalized deviation $z_{j,i}(t)$ of~\eqref{eq:demonstration_normalization} is penalized only beyond the calibrated tolerance $z_0$, through the hinge $\phi(x)=\max\{x-z_0,\,0\}$, so that deviations below the learned threshold incur no cost.
Measured directions are penalized individually, each with a tolerance $z_0\sigma_{j,i}$ set by its demonstrated spread.
Floored directions share the variance $\sigma_0^2$, so their individual eigenvectors are not unique, but the subspace they span is.
We therefore penalize the Euclidean norm of their normalized deviations, which is independent of the basis chosen within this subspace:
\begin{equation}
\begin{aligned}
    c(t)=\frac{1}{n(t)}
    \sum_{j\in\mathcal{C}}
    \Bigg(
    &\phi\Big(
        \sqrt{\textstyle\sum_{i\in\mathcal{F}_j(t)}z_{j,i}(t)^2}
    \Big) \\
    &\qquad+\sum_{i\in\mathcal{M}_j(t)}
    \phi\big(|z_{j,i}(t)|\big)
    \Bigg),
\end{aligned}
\label{eq:frame_cost}
\end{equation}
where $n(t)$ is the total number of constrained directions at frame~$t$.

\subsubsection{\textbf{Task Quality Index}}
\label{subsec:quality}
SPECTRA aggregates the frame costs over an execution into a single final quality score.
The trajectory cost $\bar{c}$ is the mean of $c(t)$ over the $N$ frames with at least one constrained direction, weighting deviations by their duration.
SPECTRA complements it with a terminal gate $\gamma\in\{0,1\}$, taken from the position conditions of each task's success check on the final object poses.
The Task Quality Index is
\begin{equation}
\boxed{
    Q=\exp\!\left(-\tfrac{1}{2}\bar{c}\right)\gamma
},
\qquad Q\in[0,1].
\label{eq:quality_index}
\end{equation}
$Q=1$ indicates that all deviations remain within $z_0$ and the execution ends in a successful configuration; lower values indicate larger average deviation beyond $z_0$, and $Q=0$ if the terminal gate fails.

For retrospective analysis, the score before gating can be traced over
time as
\begin{equation}
    Q_{\mathrm{path}}(t)
    =
    \exp\!\left(
        -\frac{1}{2N}\sum_{s\leq t}c(s)
    \right),
    \label{eq:path_quality_running}
\end{equation}
where the sum runs over the same frames as $\bar{c}$.
%
% The trace starts at $1$, is non-increasing, and drops only where a
% deviation exceeds the tolerance, so that $Q=Q_{\mathrm{path}}(T)\,\gamma$
% at the final frame $T$.

\section{Experiments}
\label{sec:experiments}

Our experiments address four research questions:
\begin{list}{}{%
    \settowidth{\labelwidth}{\textbf{RQ1}}%
    \setlength{\labelsep}{0.5em}%
    \setlength{\leftmargin}{\dimexpr\labelwidth+\labelsep\relax}%
    \setlength{\itemindent}{0pt}%
    \renewcommand{\makelabel}[1]{#1\hfil}%
}
    \item[\textbf{RQ1}] Does SPECTRA learn a meaningful and stable task representation from a small set of demonstrations?
    \item[\textbf{RQ2}] Does SPECTRA capture quality differences between learned policies, in agreement with human judgment?
    \item[\textbf{RQ3}] Does the task representation learned from simulated demonstrations transfer to scoring hardware executions?
    \item[\textbf{RQ4}] Can SPECTRA’s scores guide demonstration selection to improve policy learning?
\end{list}
We further ablate the main design choices of SPECTRA, namely the number of demonstrations $K$, the number of progress bins $M$, the pinning threshold $\sigma_{\mathrm{pin}}$, and the standard-deviation floor $\sigma_0$.

\textbf{Experimental setup:} We use ManiSkill~\cite{tao2024maniskill3} as our simulation environment and consider seven table-top manipulation tasks: \textit{StackCube}, \textit{PlaceSphere}, \textit{PegInsertion}, \textit{PlugCharger}, \textit{PokeCube}, \textit{CloseDrawer} and \textit{PlaceCan}.
Unless otherwise stated, each task-specific model uses $K=30$ demonstrations, $M=10$ progress bins per subtask and $\sigma_{\mathrm{pin}}=4\,\mathrm{mm}$.
Hardware experiments use a UR3 robotic arm, with the 6-DoF poses of objects estimated using FoundationPose~\cite{wen2024foundationpose}.
All experiments run on a single workstation with an Intel Core Ultra~9 285K CPU (24 cores), 64\,GB of RAM and an NVIDIA GeForce RTX~5090 GPU (32\,GB), and Operating System Ubuntu~22.04.

\subsection{RQ1: Does SPECTRA learn a meaningful and stable task representation from a small set of demonstrations?}
\label{sec:RQ1}

\textbf{Agreement with task geometry:}
We compute ground-truth constraint counts from each task setup: directions the scene randomizes are free, while those fixed by contact, a joint, the grasp or the solver are constrained.
We evaluate $88$ cases, one for each manipulated--reference and object--table relation, and each case has two ground-truth counts: the directions constrained throughout the sub-task and those constrained at its end.
We compare the first with the minimum learned count $|\mathcal{I}_j(l)|$ over the sub-task's $M$ samples, and the second with the learned count at its final sample, giving $176$ checks.
Of these, $166$ are correct ($94.3\%$): $82$ throughout and $84$ at the end.

\textbf{Robustness to the demonstration set:}
A reference should neither depend on which demonstrations happened to be collected nor penalize executions of the kind it was learned from.
For each task, we infer five constraint models, each on $K=30$ demonstrations drawn at random from a pool of $200$.
Across the five fits, the number of constrained directions agrees at $90\%$ of progress samples.

\textbf{Sensitivity to controlled deviations:}
We perturb motion-planned executions by tilting or rolling the object by a fixed angle at each waypoint of the carry phase, then returning it to the planned pose before placement.
\textit{CloseDrawer} has no grasped object, so the gripper is rotated instead.
For each task, we collect ten executions at each of $5^\circ$, $10^\circ$, $25^\circ$ and $45^\circ$ of deviations, together with ten clean executions.
As shown in Table~\ref{tab:perturbation}, mean $Q$ decreases monotonically with deviation magnitude in all seven tasks, and the Spearman correlation between deviation and $Q$ is strong and statistically significant in every task ($\rho$ from $-0.74$ to $-0.98$, $p<0.001$).
Small deviations within the admissible radius $z_0$ incur no cost: at $5^\circ$, \textit{StackCube} and \textit{PlaceSphere} remain within the spread of clean executions.
The drop from clean to $45^\circ$ ranges from $0.57$ on \textit{CloseDrawer} to $0.08$ on \textit{PlaceSphere}.
The small response on \textit{PlaceSphere} is consistent with the learned reference: because the sphere is symmetrical, its orientation does not affect the task, and varies across demonstrations accordingly.

\begin{table}[h]
\centering
\scriptsize
\setlength{\tabcolsep}{2.5pt}
\caption{Mean $Q$ under controlled deviations. $\rho$: Spearman's correlation between deviation and $Q$ per task.}
\label{tab:perturbation}
\begin{tabular*}{\columnwidth}{@{\extracolsep{\fill}}llcccccc@{}}
\toprule
& & \multicolumn{5}{c}{$Q$ (Ours) at commanded deviation} & \\
\cmidrule(lr){3-7}
Task & Axis & Clean & $5^\circ$ & $10^\circ$ & $25^\circ$ & $45^\circ$ & $\rho$ \\
\midrule
\textit{PlaceCan}     & tilt & 0.99 & 0.97 & 0.94 & 0.82 & 0.65 & $-0.98$ \\
\textit{CloseDrawer}  & tilt & 1.00 & 0.88 & 0.81 & 0.58 & 0.42 & $-0.97$ \\
\textit{PlugCharger}  & roll & 0.99 & 0.98 & 0.94 & 0.84 & 0.76 & $-0.94$ \\
\textit{PegInsertion} & roll & 0.99 & 0.95 & 0.90 & 0.71 & 0.56 & $-0.94$ \\
\textit{StackCube}    & tilt & 0.99 & 0.99 & 0.98 & 0.90 & 0.75 & $-0.88$ \\
\textit{PokeCube}     & roll & 1.00 & 0.98 & 0.97 & 0.91 & 0.81 & $-0.84$ \\
\textit{PlaceSphere}  & tilt & 0.99 & 0.99 & 0.99 & 0.98 & 0.92 & $-0.74$ \\
\bottomrule
\end{tabular*}
% \par\vspace{2pt}
% {\footnotesize $^{*}p<0.001$\par}
\end{table}
\vspace{-2ex} 

% \noindent\textbf{Answer to RQ1:} From $30$ demonstrations, SPECTRA learns constraint counts that match the expected counts in $94\%$ of the evaluated checks, and do not penalize new successful executions ($Q=0.99$).
% %
% Deviations from this reference lower $Q$ monotonically with their magnitude in all seven tasks, while deviations within the admissible radius remain unpenalized.

\subsection{RQ2: Does SPECTRA capture quality differences between learned policies, in agreement with human judgment?}
\label{sec:RQ2}

\begin{table}[t]
\centering
\scriptsize
\setlength{\tabcolsep}{4pt}
\caption{Mean of each metric per policy rollouts; darker blue indicates the better value within each metric.}
\label{tab:policy_means}
\begin{tabular*}{\columnwidth}{@{\extracolsep{\fill}}lccccc}
  \toprule
  Policy & $Q$ (Ours) $\uparrow$ & Qwen3-VL $\uparrow$ & SQS $\uparrow$ & $P$ (m) $\downarrow$ & $J$ (m/s$^3$) $\downarrow$ \\
  \midrule
  DP          & \cellcolor{rankone}0.75   & \cellcolor{ranktwo}0.52   & \cellcolor{rankone}0.79   & \cellcolor{rankthree}0.82 & \cellcolor{ranktwo}4.14 \\
  ACT         & \cellcolor{rankthree}0.64 & \cellcolor{rankone}0.55   & \cellcolor{ranktwo}0.78   & \cellcolor{rankone}0.63   & \cellcolor{rankone}2.13 \\
  $\pi_{0.5}$ & \cellcolor{ranktwo}0.71   & \cellcolor{rankthree}0.45 & \cellcolor{rankthree}0.74 & \cellcolor{ranktwo}0.77   & \cellcolor{rankthree}4.89 \\
  \bottomrule
\end{tabular*}
\end{table}

\begin{table}[t]
\centering
\footnotesize
\caption{Spearman correlation with mean human ratings. Each task has $15$ rollouts, $n_s$ of them successful; \textit{Overall} pools all tasks. SQS: PRM-as-a-Judge success quality score; $P$: TCP path length; $J$: mean Cartesian jerk, correlated as $-P$ and $-J$. Bold: best per row. --: undefined, as Qwen3-VL assigns all rollouts the same score.}
\label{tab:policy_quality}

\setlength{\tabcolsep}{4pt}
\begin{tabular*}{\columnwidth}{@{\extracolsep{\fill}}lccc@{}}
  \multicolumn{4}{@{}l}{\textbf{(a)} All rollouts ($N=105$)} \\
  \toprule
  Task & $Q$ (Ours) & Qwen3-VL & SQS \\
  \midrule
  \textit{CloseDrawer}  & \textbf{0.37} & 0.25 & 0.16 \\
  \textit{PlaceCan}     & \textbf{0.71} & 0.44 & 0.20 \\
  \textit{PlaceSphere}  & \textbf{0.90} & 0.37 & 0.79 \\
  \textit{PokeCube}     & \textbf{0.81} & 0.22 & 0.31 \\
  \textit{StackCube}    & \textbf{0.64} & 0.56 & 0.42 \\
  \textit{PlugCharger}  & \textbf{0.93} & 0.12 & 0.43 \\
  \textit{PegInsertion} & \textbf{0.85} & 0.40 & 0.74 \\
  \midrule
  \textit{Overall}      & \textbf{0.81} & 0.28 & 0.50 \\
  \bottomrule
\end{tabular*}

\vspace{6pt}

\setlength{\tabcolsep}{3pt}
\begin{tabular*}{\columnwidth}{@{\extracolsep{\fill}}lcccccc@{}}
  \multicolumn{7}{@{}l}{\textbf{(b)} Successful rollouts only ($N=81$)} \\
  \toprule
  Task & $n_s$ & $Q$ (Ours) & Qwen3-VL & SQS & $-P$ & $-J$ \\
  \midrule
  \textit{CloseDrawer}  & 14 & 0.22 & 0.18 & $-0.03$ & 0.29 & \textbf{0.54} \\
  \textit{PlaceCan}     & 14 & \textbf{0.65} & -- & 0.01 & 0.39 & 0.45 \\
  \textit{PlaceSphere}  & 12 & \textbf{0.89} & 0.00 & 0.63 & 0.80 & 0.46 \\
  \textit{PokeCube}     & 12 & \textbf{0.64} & $-0.11$ & 0.17 & $-0.06$ & $-0.06$ \\
  \textit{StackCube}    & 12 & 0.30 & 0.19 & 0.20 & \textbf{0.45} & 0.44 \\
  \textit{PlugCharger}  & 8  & \textbf{0.87} & $-0.62$ & $-0.38$ & $-0.02$ & 0.84 \\
  \textit{PegInsertion} & 9  & \textbf{0.44} & $-0.24$ & 0.13 & $-0.44$ & $-0.55$ \\
  \midrule
  \textit{Overall}      & 81 & \textbf{0.61} & 0.05 & 0.25 & 0.42 & 0.33 \\
  \bottomrule
\end{tabular*}
\vspace{-3ex}
\end{table}

\begin{figure}[h]
    \centering
    \includegraphics[width=0.95\linewidth]{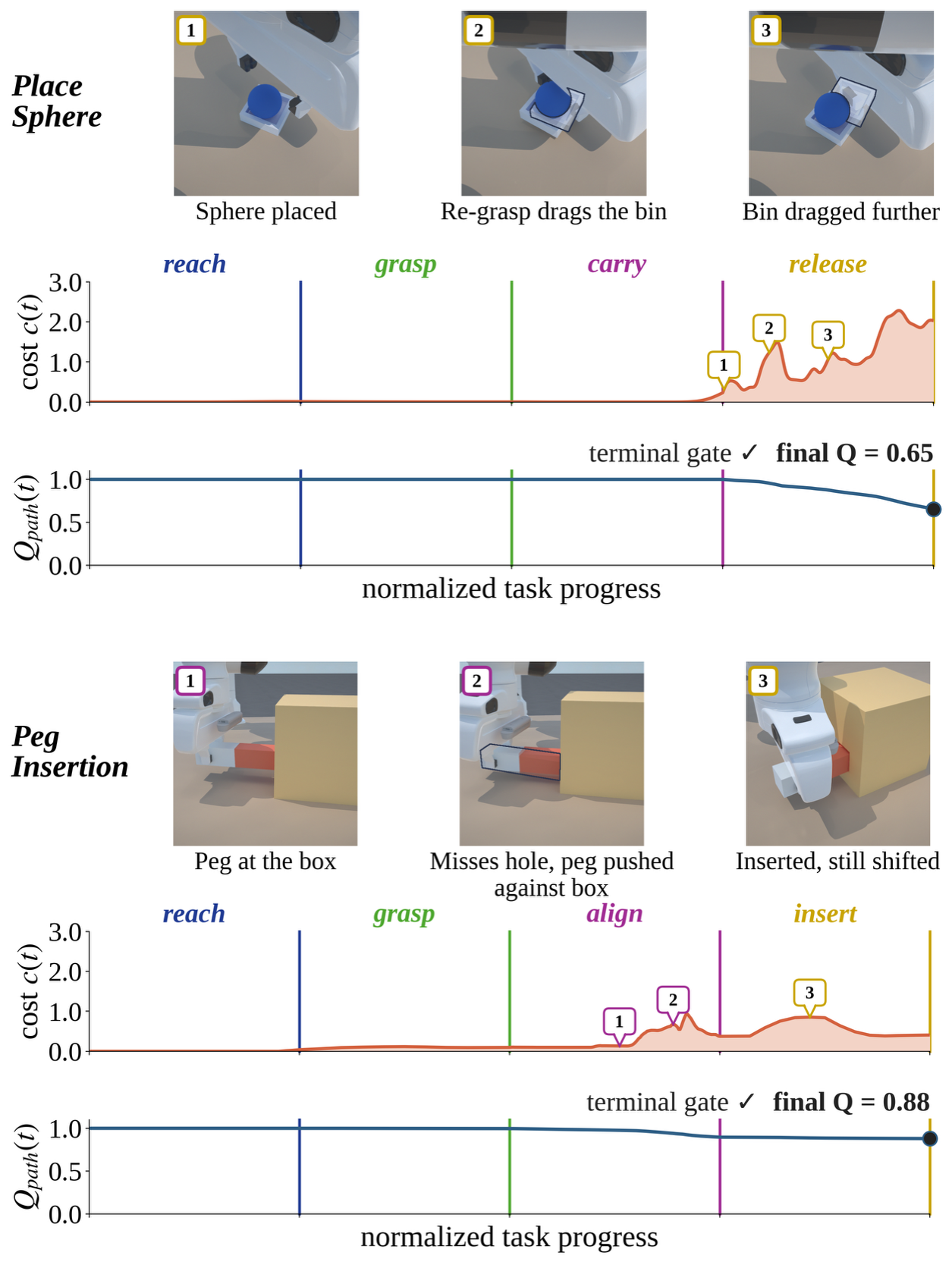} 
    \caption{Example of successful rollouts with different SPECTRA Quality score Q. Numbered keyframes mark events in the frame cost $c(t)$; $Q_{\mathrm{path}}(t)$ shows the cumulative score. The bin dragged during release and the peg pushed against the box during alignment lower their final scores.}
\label{fig:rollout_examples}
    \vspace{-2ex}
\end{figure}

We evaluate Diffusion Policy (DP)~\cite{chi2025diffusion}, ACT~\cite{zhao2023act} and $\pi_{0.5}$~\cite{black2025pi05}, each trained on $200$ ManiSkill~\cite{tao2024maniskill3} demonstrations per task.
For each policy and task, we randomly select four successful rollouts and one failure; where a policy has fewer than four successes, the remaining rollouts are failures, and where it has no failures, all five are successes, giving $105$ executions: $81$ successful and $24$ failed.
Fifteen raters score each execution from $1$ (failure) to $5$ (perfect), with policy identities concealed and task and rollout order randomized for each rater.
Inter-rater agreement is ICC$(2,1)=0.82$ for a single rater and ICC$(2,k)=0.99$ for the mean of all raters, with ordinal Krippendorff's $\alpha=0.76$; on successful rollouts only, these are $0.58$, $0.95$ and $0.52$.
Judging quality among successful executions is thus harder for individual raters, but the mean rating remains reliable, and we use it as the reference.

We compare SPECTRA Quality Index $Q$ with four baselines.
PRM-as-a-Judge's Success Quality Score (SQS)~\cite{liu2026prmasajudge15toolkitrobot} tracks learned task progress, with its reward model fine-tuned with LoRA on $50$ demonstrations per task.
As a vision-language judge, Qwen3-VL-32B is given the same task description and quality criteria as the human raters and asked to score each video with a number $s$ between $1$ and $5$, which we map to $[0,1]$ as $(s-1)/4$ for comparison with $Q$ and SQS.
We also include two kinematic indicators commonly reported alongside success rate~\cite{valle2025evaluating,wang2025roboeval}: the Cartesian path length of the tool center point, $P$, and its mean Cartesian jerk, $J$.
Metrics disagree on which policy performs best (Table~\ref{tab:policy_means}): $Q$ and SQS rank DP highest, whereas Qwen3-VL, $P$ and $J$ favor ACT, which $Q$ ranks lowest.
We therefore assess agreement with human judgment at the level of individual rollouts, using Spearman's rank correlation (SRCC) with the mean human rating.
$Q$, SQS and Qwen3-VL account for task success and are compared over all rollouts; since $P$ and $J$ do not, all metrics are also compared over successful rollouts only.

\textbf{All rollouts:}
Over all $105$ rollouts, $Q$ achieves stronger pooled agreement with human ratings ($\rho=0.81$) than SQS ($0.50$) and Qwen3-VL ($0.28$), and the highest correlation within each of the seven tasks (Table~\ref{tab:policy_quality}).

\textbf{Successful executions:}
On the $81$ successful rollouts, $Q$ achieves the highest pooled agreement ($\rho=0.61$), followed by $-P$ ($0.42$), $-J$ ($0.33$), SQS ($0.25$) and Qwen3-VL ($0.05$), and agrees best on five of the seven tasks: \textit{PlaceSphere} ($0.89$), \textit{PlugCharger} ($0.87$), \textit{PlaceCan} ($0.65$), \textit{PokeCube} ($0.64$) and \textit{PegInsertion} ($0.44$).
On \textit{PegInsertion}, path length and jerk are even negatively correlated with the ratings ($-0.44$ and $-0.55$): the executions raters judged best are not the shortest or smoothest, but the ones that align the peg precisely with the hole.
Qwen3-VL, which agrees over all rollouts mainly by detecting failures, is negatively correlated among successes on \textit{PlugCharger} ($-0.62$) and \textit{PegInsertion} ($-0.24$): the small misalignments separating these executions are difficult to assess from video, whereas SPECTRA measures them directly in the relative object poses.
The kinematic metrics agree best on the remaining two tasks, capturing complementary aspects of quality.
On \textit{CloseDrawer}, where fewer geometric relations constrain the task and DP and ACT achieve similar $Q$, raters separate them by smoothness: every ACT rollout has lower jerk than any DP rollout, and $-J$ agrees best among successes ($0.54$).
On \textit{StackCube}, path length agrees best ($0.45$).

Fig.~\ref{fig:rollout_examples} shows two successful executions with lower $Q$.
In \textit{PlaceSphere}, the score decreases because the gripper drags the bin while releasing the sphere; in \textit{PegInsertion}, SPECTRA captures a subtle misalignment, as the peg misses the hole and is pushed against the box before insertion.
Neither deviation slows task progress or changes the robot's motion; both disturb the relation between objects, which SQS, path length and jerk do not measure.

% \vspace{2ex}
% \noindent\textbf{Answer to RQ2:} SPECTRA agrees with human ratings more closely than all baselines, both over all rollouts ($\rho=0.77$, versus at most $0.50$) and among successful executions ($\rho=0.54$, versus at most $0.42$), most clearly on tasks requiring precise relative placement.
% %
% Robot-motion metrics capture complementary aspects of quality, such as smoothness, on tasks where geometric precision varies little.

\subsection{RQ3: Does the task representation learned from simulated demonstrations transfer to scoring hardware executions?}
\label{sec:RQ3}

We first evaluate controlled, motion-planned executions, which are expected to achieve high geometric quality and provide a baseline for assessing transfer to hardware.
We apply the task representation learned from ManiSkill oracle demonstrations, to \textit{StackCube} and \textit{PlaceCan} executions planned with DrakeIK~\cite{drakeik} on a physical UR3 arm.
The representation transfers with little loss: across ten executions per task, the mean scores are $Q=0.95\pm0.01$ for \textit{StackCube} and $Q=0.93\pm0.01$ for \textit{PlaceCan}, close to $Q=0.99$ of unseen successful executions in simulation, and consistent across executions.
The remaining cost arises from the gripper--object relations, as physical grasps differ from those in the oracle demonstrations: the cube is held $4$--$7\,\mathrm{mm}$ off-centre, and the can $1$--$6\,\mathrm{mm}$ off-centre, whereas the oracle centres it to within $0.2\,\mathrm{mm}$.

\begin{figure}[h!]
    \centering
    \includegraphics[width=0.9\linewidth]{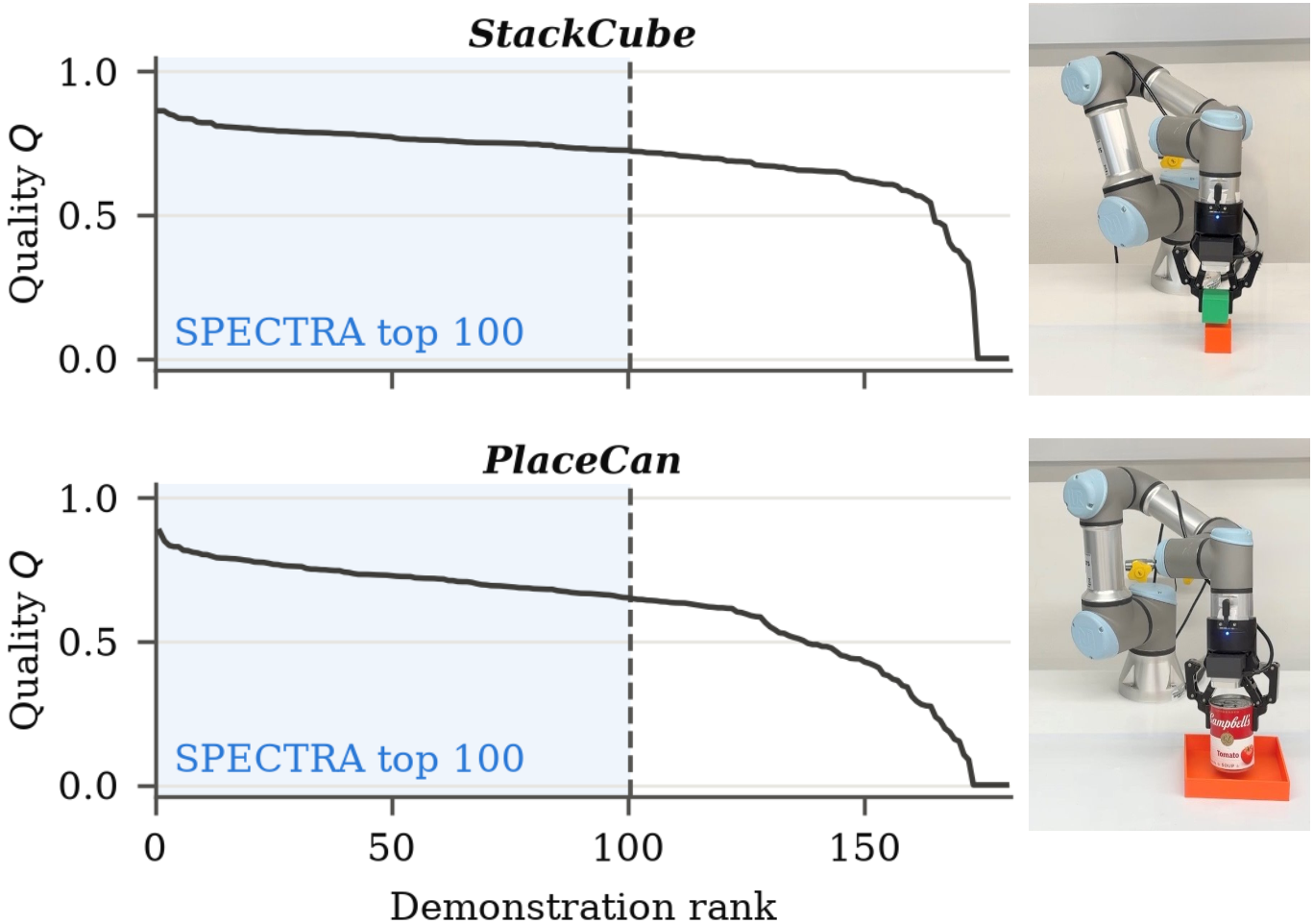} 
    \caption{SPECTRA quality index Q of the 180 teleoperated demonstrations per task, sorted by rank.}
    \label{fig:RQ3}
\end{figure}

Teleoperated executions, on the other hand, may vary more in grasp configuration and object alignment, and are therefore expected to receive lower and more variable scores.
We score the 180 teleoperated demonstrations per task recorded for RQ4, all judged successful by visual inspection, with the same reference: they score $Q=0.61\pm0.28$ on \textit{StackCube} and $Q=0.62\pm0.18$ on \textit{PlaceCan} (Fig.~\ref{fig:RQ3}).
These scores include demonstrations assigned $Q=0$ by the terminal gate: five on \textit{StackCube}, where the top cube is released misaligned with the bottom cube, and eight on \textit{PlaceCan}, where the can is released off-centre against the tray wall.
In both cases, the placement looks acceptable on visual inspection but is not geometrically precise enough to meet the terminal position criteria, showing that SPECTRA flags imprecision that visual judgment overlooks.

\vspace{-2ex}
\vspace{-3ex}

\subsection{RQ4: Can SPECTRA's scores guide demonstration selection to improve policy learning?}
\label{sec:RQ4}

We conduct an exploratory study of whether SPECTRA's scores can support demonstration selection for Diffusion Policy training, using the teleoperated pools scored in Sec.~\ref{sec:RQ3}.
For each task, we train one policy on the 100 highest-scoring demonstrations and another on 100 selected at random from the same pool, each with one training seed and evaluated over ten executions on the UR3 arm.
Policies trained on SPECTRA-selected demonstrations succeed in 9/10 executions on both tasks, compared with 5/10 on \textit{StackCube} and 6/10 on \textit{PlaceCan} for random selection, suggesting that SPECTRA can support demonstration selection.

\subsection{Ablations}
\label{sec:ablations}
 
We ablate the number of demonstrations $K$, the number of progress bins $M$, the pinning threshold $\sigma_{\mathrm{pin}}$, and the standard-deviation floor $\sigma_0$ (Table~\ref{tab:ablation}).
%
% Each measure tests one property of the reference: Rec.\ its constraint recovery (Sec.~\ref{sec:RQ1}), $\bar Q_{\mathrm{clean}}$ whether clean executions remain unpenalized, and $\Delta Q_d=\bar Q_{\mathrm{clean}}-\bar Q_d$ whether a commanded deviation of $d$ degrees (Table~\ref{tab:perturbation}) is detected.
 
\begin{table}[t]
\centering
\footnotesize
\setlength{\tabcolsep}{4pt}
\renewcommand{\arraystretch}{1.0}
\caption{Ablations, one parameter at a time with the others at their defaults ($^\dagger$). Rec.: constraint recovery. $\bar Q_{\mathrm{clean}}$: mean $Q$ on unseen clean executions. $\Delta Q_{d}$: mean decrease in $Q$ under a $d^\circ$ deviation. $\sigma_{\mathrm{pin}}$, $\sigma_0$ in mm.}
\label{tab:ablation}
\begin{tabular*}{\columnwidth}{@{\extracolsep{\fill}}lcccc@{}}
\toprule
Setting & Rec.\,(\%) & $\bar Q_{\mathrm{clean}}$ & $\Delta Q_{10}$ & $\Delta Q_{45}$ \\
\midrule
\multicolumn{5}{@{}l}{\textit{Demonstration budget} $K$} \\
5             & 82.4 & 0.91 & 0.088 & 0.347 \\
10            & 92.6 & 0.98 & 0.075 & 0.316 \\
30$^\dagger$  & 94.3 & 0.99 & 0.055 & 0.294 \\
70            & 96.0 & 0.99 & 0.052 & 0.287 \\
\midrule
\multicolumn{5}{@{}l}{\textit{Within-phase resolution} $M$} \\
1             & 88.6 & 0.91 & 0.017 & 0.135 \\
5             & 93.8 & 0.99 & 0.040 & 0.245 \\
10$^\dagger$  & 94.3 & 0.99 & 0.055 & 0.294 \\
15            & 94.9 & 0.99 & 0.058 & 0.301 \\
\midrule
\multicolumn{5}{@{}l}{\textit{Pinning threshold} $\sigma_{\mathrm{pin}}$} \\
4$^\dagger$   & 94.3 & 0.99 & 0.055 & 0.294 \\
7             & 85.8 & 0.99 & 0.065 & 0.309 \\
9             & 75.0 & 0.99 & 0.067 & 0.312 \\
$\infty$ (all six) & 5.7 & 0.99 & 0.049 & 0.239 \\
\midrule
\multicolumn{5}{@{}l}{\textit{Standard-deviation floor} $\sigma_0$} \\
1             & 94.3 & 0.99 & 0.135 & 0.569 \\
2             & 94.3 & 0.99 & 0.075 & 0.394 \\
3$^\dagger$   & 94.3 & 0.99 & 0.055 & 0.294 \\
3.5           & 94.3 & 0.99 & 0.046 & 0.255 \\
\bottomrule
\end{tabular*}
\end{table}

\textbf{Demonstration budget:}
Because deviations are measured from their mean, $K$ demonstrations span at most $K-1$ of the six tangent directions. At $K=5$, at least two directions thus show no variation simply for lack of samples; flooring sets them to $\sigma_0<\sigma_{\mathrm{pin}}$, so they are pinned regardless of task geometry.
Recovery drops to $82.4\%$ and clean executions are penalized along these spuriously pinned directions ($\bar Q_{\mathrm{clean}}=0.91$), so the larger $\Delta Q$ reflects over-constraint rather than better detection. 
From $K=30$, all three measures stabilize.
 
\textbf{Within-phase resolution.}
With $M=1$, each sub-task is described by a single nominal pose and covariance, so a relation that changes within the sub-task, such as an object being lowered onto its target, is compared against its average over the whole sub-task.
Ordinary motion then appears as deviation ($\bar Q_{\mathrm{clean}}=0.91$), while the spread pooled over the sub-task absorbs genuine deviations: $\Delta Q_{45}$ falls to $0.135$, less than half its default value.
Both measures recover by $M=5$--$10$, showing that progress-dependent modeling is needed both to leave clean executions unpenalized and to detect deviations.
 
\textbf{Constraint identification:} Since each direction is normalized by its own spread, clean executions remain unpenalized for every threshold ($\bar Q_{\mathrm{clean}}=0.99$); $\sigma_{\mathrm{pin}}$ instead determines which directions enter the cost.
Raising it to $9\,\mathrm{mm}$ pins directions the task does not constrain, lowering recovery to $75\%$ with little change in sensitivity.
Pinning all six ($\sigma_{\mathrm{pin}}=\infty$) removes the constraint structure (Rec.\ $5.7\%$) and weakens detection ($\Delta Q_{45}$: $0.294\rightarrow0.239$), as free directions add terms to $n(t)$ but little penalty, diluting genuine deviations.
We use $4\,\mathrm{mm}$, which gives the highest recovery.

\textbf{Standard-deviation floor:}
Since $\sigma_0<\sigma_{\mathrm{pin}}$, the floor does not affect recovery but sets the strictness on tightly regulated directions: at $1\,\mathrm{mm}$, $\Delta Q_{45}$ nearly doubles ($0.569$), penalizing millimetre-scale departures comparable to hardware pose-estimation accuracy.
We use $3\,\mathrm{mm}$, which tolerates such departures while preserving sensitivity to larger deviations, which decreases further at $3.5\,\mathrm{mm}$ ($\Delta Q_{45}=0.255$).

% \vspace{2ex}
% \noindent\textbf{Summary:} Performance largely stabilizes around $30$ demonstrations and $5$--$10$ progress bins per sub-task. Clean-execution scores remain stable across the tested pinning thresholds and floors, while sensitivity to deviations changes with the floor; penalizing all six directions instead of the identified constraints reduces sensitivity to large deviations by about a fifth.
\vspace{-1ex}
\section{Discussion and limitations}
\label{sec:discussion}

$Q$ measures adherence to the demonstrated reference rather than an absolute notion of quality, so a valid strategy absent from the demonstrations would be scored as deviating.
However, the reference can encode preferences beyond task success: for example, if the \textit{PlaceCan} demonstrations consistently placed the can with its label facing the user, that orientation would become constrained, tailoring $Q$ to a user or application through the demonstrations alone.
Demonstrations should therefore reflect intended range of admissible executions, which simulation makes inexpensive to control.

Policy evaluation rarely measures the geometric precision with which objects interact.
Our results suggest it is distinct from motion quality: $Q$ agrees best with human ratings on tasks requiring precise relative placement, whereas jerk agrees best on \textit{CloseDrawer}, where raters judge the smoothness of the robot's motion.
Unlike kinematic metrics, $Q$ is bounded in $[0,1]$ and expressed relative to each task's admissible variation, making it comparable across tasks.
It is also diagnosable: the frame cost shows when an execution deviates and which relation is affected.

SPECTRA depends on correct sub-task segmentation and, on hardware, on accurate pose estimation.
For now, it applies only to rigid bodies and relies on oracle demonstrations scripted in simulation, which limits it to simple tasks whose solutions can be scripted.

\vspace{-1ex}
\section{Conclusion}
\label{sec:conclusion}
We introduced SPECTRA, an object-centric framework for evaluating geometric execution quality in rigid-body manipulation.
From $30$ oracle demonstrations generated in simulation, SPECTRA learns which relative-pose directions are constrained at each stage of a task and how much variation each admits, recovering the expected constraint counts in $94\%$ of cases.
Scoring unseen executions by their deviations beyond these tolerances, it agrees with human ratings more closely than all baselines, both over all rollouts ($\rho=0.81$) and among successful executions ($\rho=0.61$).
The task reference learned in simulation transfers to scoring hardware executions, and an exploratory study suggests that its scores can guide demonstration selection.

In future work, we will use egocentric human videos as oracle demonstrations to apply SPECTRA to more complex tasks.
We also plan to use SPECTRA beyond evaluation, both for curating demonstration datasets and as a dense reward for policy learning.

\bibliographystyle{IEEEtran}
\bibliography{references}

\end{document}